\pdfoutput=1

\documentclass[11pt]{article}

\usepackage[]{acl}

\usepackage{times}
\usepackage{latexsym}

\usepackage[T1]{fontenc}

\usepackage[utf8]{inputenc}

\usepackage{microtype}
\usepackage{tabularx}
\usepackage{multirow}
\usepackage{amsmath}
\usepackage{amssymb}
\usepackage{amsthm}
\usepackage{pifont}
\usepackage{xcolor} 
\usepackage{hyperref}       
\usepackage{url}            
\usepackage{booktabs}       
\usepackage{amsfonts}       
\usepackage{nicefrac}       
\usepackage{microtype}      
\usepackage{xcolor}         

\usepackage{amsmath}
\usepackage{amssymb}

\usepackage{mathtools}
\usepackage{amsthm}
\usepackage{multirow}
\usepackage{bbm}
\usepackage{boldline}
\usepackage{amssymb}

\usepackage{algorithm}
\usepackage{algorithmic}

\usepackage{mathtools}

\title{The steerability of large language models toward data-driven personas}

\newcommand\equalauthor{\thanks{* These authors contributed equally to this work}}

\author{{\bf Junyi Li$^1$, Charith Peris$^2$\equalauthor, Ninareh Mehrabi$^{2*}$, Palash Goyal$^2$, Kai-Wei Chang$^2$},\\  {\bf Aram Galstyan$^2$, Richard Zemel$^2$ and Rahul Gupta$^2$}\\
$^1$University of Maryland, College Park\\
$^2$Amazon\\
junyili.ai@gmail.com\\
\{perisc, ninarehm, palashg, kaiwec, argalsty, rzemel, gupra\}@amazon.com}

\begin{document}
\maketitle
\begin{abstract}

Large language models (LLMs) are known to generate biased responses where the opinions of certain groups and populations are underrepresented. Here, we present a novel approach to achieve controllable generation of specific viewpoints using LLMs, that can be leveraged to produce multiple perspectives and to reflect the diverse opinions. Moving beyond the traditional reliance on demographics like age, gender, or party affiliation, we introduce a data-driven notion of {\em persona} grounded in collaborative filtering, which is defined as either a single individual or a cohort of individuals manifesting similar views across specific inquiries. As individuals in the same demographic group may have different personas, our data-driven persona definition allows for a more nuanced understanding of different (latent) social groups present in the population. In addition to this, we also explore an efficient method to steer LLMs toward the personas that we define. We show that our data-driven personas significantly enhance model steerability, with improvements of between $57\%-77\%$ over our best performing baselines.

\end{abstract}

\begin{figure*}[t]
    \centering
    \includegraphics[width=0.95\linewidth]{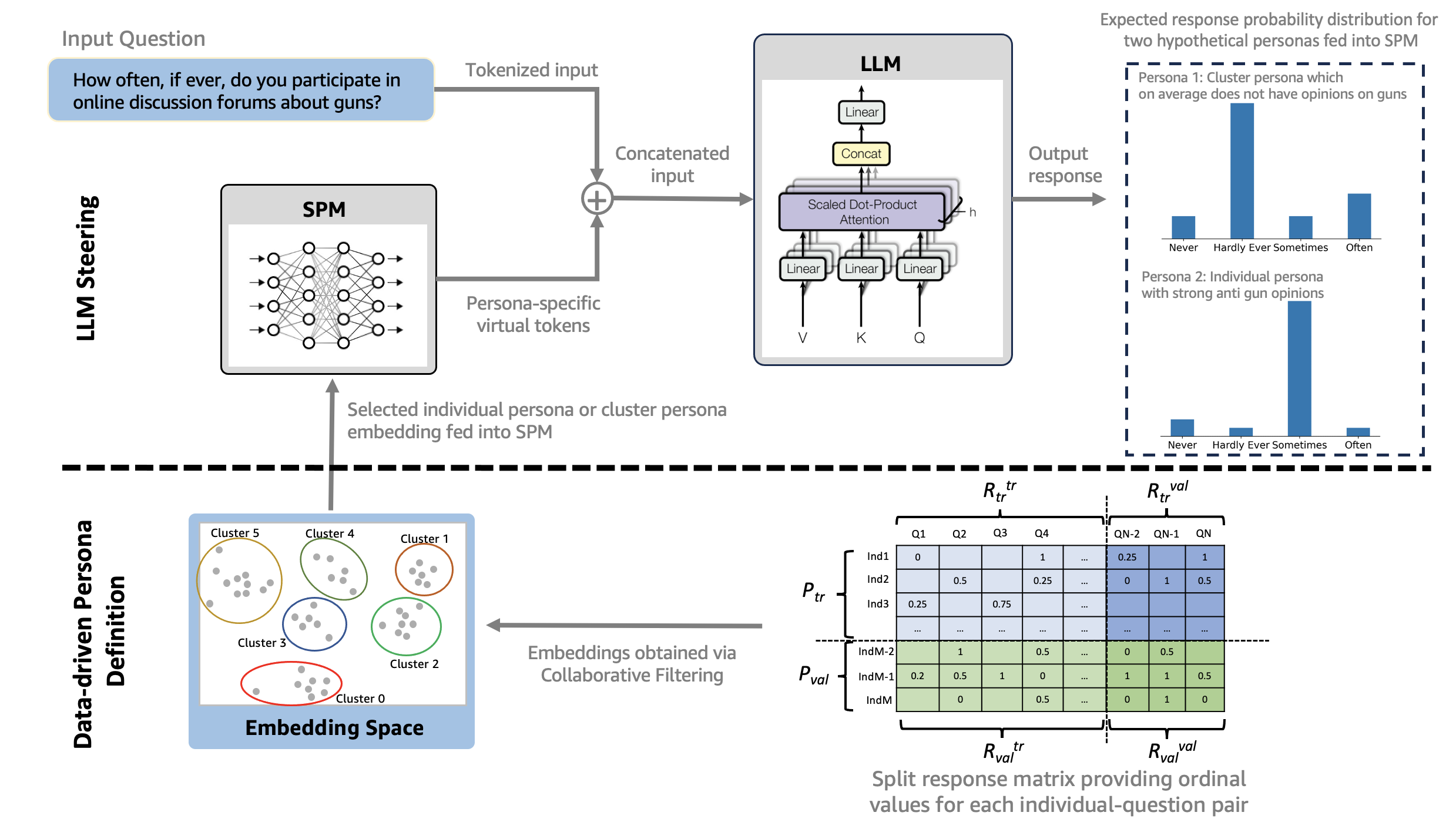}
    \caption{A schematic of our framework for steering LLMs toward data-driven personas. The bottom-half 
    illustrates the formation of 
    data-driven personas, 
    and the top-half illustrates 
    LLM steering. A persona is defined by
    generating individual embeddings via collaborative filtering. 
    The persona can be a single individual embedding (grey dots) or the centroid of a group of embeddings, referred to as a cluster persona (denoted by the circled clusters). To steer the LLM we pass an embedding to a soft-prompting model (SPM), which maps the embedding to a set of persona-specific virtual tokens. Finally we prepend these virtual tokens to the tokenized input sequence and pass this into the LLM to obtain a persona-specific response. }
    \label{fig:steer-structure}
\end{figure*}

\section{Introduction}
\label{sec:intro}
In the recent past, Large Language Models (LLMs; ~\citealt{brown2020language, ouyang2022training, openai2023gpt, touvron2023llama}) have shown exceptional generation capabilities across a range of tasks~\cite{zhao2023survey} that have led to their adoption in applications pertaining to multiple high-stakes fields such as healthcare~\cite{singhal2023large}, education~\cite{tan2023towards} and finance~\cite{wu2023bloomberggpt}. Given this context, it is of utmost importance to avoid biases towards specific under-represented populations, and leverage LLMs in a way that enables generation across a broad spectrum of viewpoints in a balanced way.


In practice, LLMs have been shown to generate responses that represent a wide range of opinions, with tendencies to over-represent the opinions of certain populations while under-representing those of others. For example, ~\citet{santurkar2023whose} showed that LLMs under-represent the opinions of individuals aged 65 and over, Mormons, and the widowed, which constitute significant portions of the US population. The reason for this is that the typical fine-tuning of an LLM, done across datasets at hand, leads to a model gaining a randomized viewpoint based on the nature of the dataset. 

Instead of fine-tuning towards such a randomized viewpoint, it is desirable to enable LLMs to have {\it controllable generation} that can be steered towards specific viewpoints. This can be leveraged to produce multiple perspectives and in turn encourage diversity through the curated inclusion of a broad spectrum of viewpoints. Such a diverse set of perspectives, enabled via controllable generation, can be extremely helpful in diminishing polarization and preventing the marginalization of the voices of minority groups. 

Prior work has attempted to control LLM generation by aligning models toward demographic groups that are defined based on features such as age, gender, political party affiliation~\cite{santurkar2023whose, hwang2023aligning, simmons2022moral, jiang2022communitylm, feng2023pretraining, salewski2023context}. However, we argue that a simple group definition based on demographic features might not be sufficient to represent the nuances of the underlying different social groups present in a given population.

In this paper, we present a new approach to achieve controllable generation of specific viewpoints using LLMs. We hypothesize that for a given dataset comprising of responses provided by a population of individuals to a set of questions, there is a space of differing characteristic opinions and beliefs. We map this space to an embedding space. We then propose the notion of {\it personas}, as examples from within this personality space. We use the term {\it persona} to refer to a portion of the embedding space which represents similar opinions and beliefs over a set of questions in our dataset of choice. This could range from a single individual embedding, referred to as an {\it individual persona}, to a group of individual embeddings represented by their centroid in embedding space, referred to as a {\it cluster persona}. To create this mapping we take a data-driven approach using collaborative-filtering to embed individuals into a continuous vector space which, to our knowledge, has not been done before. Compared to the use of traditional demographic traits, our definition of 
personas allows for nuanced understanding of different social groups in the population and makes the notion of steerability more meaningful.

Given these data-driven personas, we then propose an efficient algorithm to steer LLMs towards both an individual and a cluster persona. In particular, we learn a soft-prompting model (SPM) which maps the embedding of a persona to a set of virtual tokens. These virtual tokens are prepended before tokens mapping to the actual input text, to steer the responses of the LLMs. 

We conduct a number of experiments using the OpinionQA dataset~\cite{santurkar2023whose} to evaluate the efficacy of our persona definition and the LLM steering algorithm. Our experiments show that LLMs steered via personas can align to the opinions of individuals and groups better than baseline methods. In particular, the personas defined using our approach result in more accurate prediction of opinions when compared to the use of personas based on demographic-traits.

We summarize our \textbf{contributions} as follows:  
\begin{itemize}
    \item We propose a hitherto unexplored paradigm in steerable generation where, instead of pre-defined demographic features, we use a data-driven notion of a persona to modulate the generation process. In particular, we use collaborative-filtering to embed opinions of individuals within a dataset into a continuous vector space (individual personas), and then cluster groups of individuals with similar embeddings (cluster personas).
    \item We propose a simple model and an efficient algorithm to align LLMs with these personas (both individuals and clusters).
    \item Finally, we benchmark our approach using a selected set of LLMs.
\end{itemize}

\section{Related Work}
\label{sec:relwork}
\noindent\textbf{Steering LLMs:} In the recent past, several studies have investigated opinions expressed by LLMs~\cite{santurkar2023whose, durmus2023towards, scherrer2023evaluating, santy2023nlpositionality}. For example, \citet{santurkar2023whose} curated the OpinionQA dataset based on public opinion polls and evaluated the alignment of LLM opinions with 60 U.S. demographic groups. They found a substantial misalignment between the views reflected by current LLMs and those of U.S. demographic groups (we mention a cross section of these in Section~\ref{sec:intro}). In another case, \citet{scherrer2023evaluating} designed a survey comprising both high-ambiguity and low-ambiguity moral scenarios to study the moral beliefs encoded in different LLMs. 

A variety of methods have also been proposed to steer the generation of LLMs toward specific opinions~\cite{hwang2023aligning, simmons2022moral, durmus2023towards, argyle2023out, deshpande2023toxicity, jiang2022communitylm, feng2023pretraining}. For instance, \citet{santurkar2023whose} added demographic information (such as Democratic or Republican affiliations) to prompts in order to steer the opinions of LLMs toward those groups. \citet{hwang2023aligning} worked towards aligning LLMs with individuals by incorporating their past opinions. \citet{simmons2022moral} constructed four prompts based on combinations of {\it liberal} and {\it conservative} political identities, as well as moral and immoral stances, to steer the moral beliefs of LLMs. \citet{jiang2022communitylm} fine-tuned LLMs using tweets authored by Democrats and Republicans to improve the model's alignment with the opinions of these groups. 

In contrast to this prior work, which steer LLMs based on demographic traits, our approach using data-driven personas allows for a more expressive and nuanced understanding of the different social groups present in the overall population. This, in turn, enhances the applicability of model steerability.

\noindent\textbf{Parameter-Efficient Fine Tuning:} Given the rapid increase in size of LLMs, fine-tuning the full model has become costly, giving rise to various methods that only tune a portion of model parameters~\cite{hu2021lora, lester2021power, li2021prefix, liu2021gpt, liu2022few, zhang2023llama, Ozdayi2023ControllingTE}. In one such approach, ~\citet{hu2021lora} utlize low-rank decomposition to represent weight updates with two smaller matrices (called update matrices), and then only fine-tune the decomposition matrix. Several approaches have been proposed where virtual tokens can be tuned to align a model towards a specific task. \citet{lester2021power} trains a sequence of virtual tokens that have been prepended to the input whereas ~\citet{li2021prefix} optimize a task-specific vector of virtual tokens called the prefix. ~\citet{zhang2023llama} address the instability at the initial training phase using a learnable gate to control the effect of virtual tokens over generation. 

\section{Framework}
In this section, we propose the notion of data-driven personas (Section~\ref{sec:CF}) and an efficient algorithm for steering an LLM towards them (Section~\ref{sec:steer}). An illustration of our process is shown in Figure~\ref{fig:steer-structure}.

\subsection{Data-driven persona definition}\label{sec:CF}
Instead of relying on demographic information such as age, gender or party affiliation, we use collaborative filtering to embed all individuals into a continuous vector space based on their opinions. Then a \textit{persona} is defined as a portion of the embedding space which represents similar opinions and beliefs. In particular, an \textit{individual persona} is represented by a single individual embedding, while a \textit{cluster persona} is represented by the centroid of a cluster of individuals.

Assume we have a set of questions $\mathcal{Q}$, consisting of multiple-choice questions that feature options with an ordinal structure. Furthermore, assume that we also have responses for these questions, given by a set of individuals $\mathcal{P}$. We represent these responses as a matrix $R$. If individual $i$ responds to question $j$, the element $r_{i,j} \in [0,1]$ represents the individual $i$'s response, where the responses are mapped to the interval $[0,1]$. If no response is given, $r_{i,j}$ is set to null. $\mathcal{R} = \{(i,j), \text{where } r_{i,j} \text{ is not null}, i \in \mathcal{P}, j \in \mathcal{Q}\}$ denotes the full set of responses. We utilize collaborative filtering (CF) to learn a continuous representation for each individual. More specifically, we denote the individual embeddings as $\{u_i \in \mathbb{R}^d, i\in\mathcal{P}\}$ and the question embeddings as $\{q_j \in \mathbb{R}^d, j\in\mathcal{Q}\}$ and optimize the following objective:
\begin{align}
    \underset{\parbox{3cm}{\centering $\{u_i \in \mathbb{R}^d, i\in\mathcal{P}\}$ \\ $\{q_j \in \mathbb{R}^d, j\in\mathcal{Q}\}$}}{min}\; 
    \sum_{(i,j) \in \mathcal{R}}
    \mathcal{L}(\langle u_i, q_j \rangle, r_{i,j})
\label{eq:CF}
\end{align}
where $\mathcal{L}$ is a loss objective, \emph{e.g.} the mean square error; $\langle\cdot\rangle$ denotes the inner product.

By optimizing the objective in~\eqref{eq:CF}, we obtain the converged embeddings $\{u_i \in \mathbb{R}^d, i\in\mathcal{P}\}$ that encode important information about individual opinions.  We explore steerability towards individual embeddings (\textit{individual personas}), as well as towards clusters of individuals with similar opinions within the overall population (\textit{cluster personas}). 


\subsection{Steering LLMs towards data-driven personas}\label{sec:steer}

For steering LLMs towards the personas defined in Section~\ref{sec:CF}, we draw from the {\it prefix-tuning} technique~\cite{li2021prefix} which prepends and tunes a set of prefix-vectors to each model layer that yields improvements in model generation towards specific tasks~\cite{lester2021power, Ozdayi2023ControllingTE, li2021prefix, liu2021gpt, liu2022few}. In our approach, we use a separate model that we refer to as the soft-prompting model (SPM), to map a given embedding (of a persona) from our continuous vector space, to a set of prefix-vectors. These are prepended to the tokenized input (in this case the tokenized question) and enables the steering of the generation process towards that specific persona. A schematic of this process if shown in the top-half of Figure~\ref{fig:steer-structure}.



Unlike vanilla prefix-tuning which trains a single set of virtual tokens for a given task, our method uses a single SPM to generate a sets of virtual tokens for all personas in our dataset. This is beneficial in that it is cost effective and also performant, since personas with proximate embeddings often reflect analogous opinions.


We train the SPM 
by optimizing the following objective:
\begin{align}\label{eq:1}
    \underset{\theta \in \Theta}{min}\; \sum_{(i,j) \in \mathcal{R}}\mathcal{L}(\text{LLM}(f(u_i; \theta), \text{Q}_j), \text{R}_{i,j})
\end{align}
where $\mathcal{R}$ is the set of responses, $Q_j$ is the tokenized representation of question $j$, $R_{i,j}$ is the tokenized representation of the response of individual $i$ to question $j$, $u_i$ is the embedding of individual $i$, $f(\cdot; \theta)$ denotes the SPM parameterized by $\theta$, and $\mathcal{L}$ denotes the loss objective, which in this case is cross entropy loss. During SPM training, the LLM weights are frozen and we only use individual embeddings, then during inference, we fix the weights of the SPM and steer the LLM generation towards the opinions of a specific persona by feeding its embedding as the input to the SPM.

To test the robustness of our technique to other prompting methods, we also test if our method works when using prompt-tuning \citep{lester2021power} (as opposed to prefix-tuning \citealt{li2021prefix}). For details, please see Appendix~\ref{sec:more_results}.

\section{Experimental results}
In this section we present our data, experimental setup and empirical results on data-driven persona definition and LLM steerability.

\subsection{Dataset Details}

We use the OpinionQA dataset~\cite{santurkar2023whose} which includes opinions of a diverse set of individuals over a wide range of different topics, for our work. The OpinionQA dataset is curated from 15 American Trends Panel polls, and encompasses responses from 18,339 participants to 1,476 multiple-choice questions across 23 different topics. A full list of topics covered by OpinionQA is provided in Appendix~\ref{sec:more_data}. The demographic information for each participant (i.e., Race, Ideology, Education etc.) are included in the dataset and we also include this information in Appendix~\ref{sec:more_data}. 

The response options for each question in the OpinionQA dataset is presented as ordinals, and we map each option to a numerical value. For example, for one set of response options {\it Worry a lot}, {\it Worry a little}, and {\it Not worry at all}, we assign values of 0, 0.5, and 1, respectively. We then use the numerical values of the responses as labels in equation~\eqref{eq:CF} to learn the individual embeddings.

\subsection{Analysis of cluster personas}
Here, we present an analysis of the cluster personas obtained from our data-driven approach described in Section~\ref{sec:CF}.

{\bf Cluster Definition:} We employ the matrix factorization approach (Eq.~\eqref{eq:CF}) to perform collaborative filtering. We embed individuals and questions into a 16-dimensional space and employ mean square error as our training objective. The learned individual embeddings are referred to as \textit{individual personas}. As for \textit{cluster personas}, we cluster individual embeddings using K-Means clustering. We run K-Means with varying numbers of clusters, denoted as $k$, and for each value of $k$, we replace individual embeddings with the centroids of each cluster to evaluate Eq.~\eqref{eq:CF}. This measures how well the cluster centroids can represent individuals and we use it as the criteria to select $k$. Specifically, we utilize the {\it elbow} heuristic~\cite{bishop2006pattern} to choose $k=6$, and we label these clusters as Cluster-X, where X ranges from 0 to 5, which represent the six cluster personas that we choose. 

For each cluster persona, we present the demographic composition across 13 different traits. We also detail the characteristics of opinions within each cluster, specifying questions for which a cluster disagrees with the overall population and questions where the clusters disagree with each other.

\begin{figure}[ht]
    \centering
    \includegraphics[trim={1cm 0.75cm 1cm 0.75cm}, width=0.8\columnwidth]{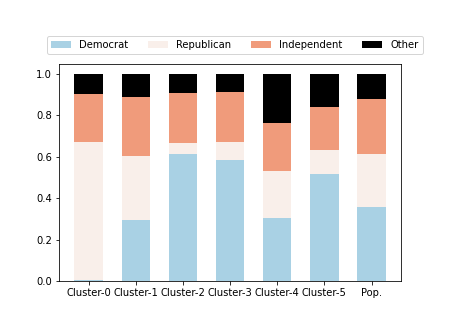}
    \includegraphics[trim={1.3cm 0.75cm 1cm 0.75cm}, width=0.75\columnwidth]{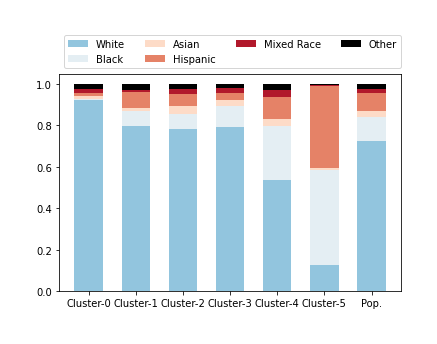}
    \includegraphics[trim={1cm 0.75cm 1cm 0.75cm}, width=0.8\columnwidth]{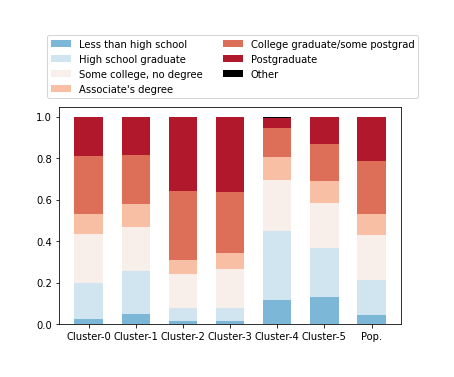}
    \caption{The Political Party, Race and Education composition (from top to bottom) of clusters and overall population.}
    \label{fig:demo_dist}
\end{figure}

\begin{table*}[ht]
\small
\centering
\caption{Comparison of responses between Cluster 0 and the overall population}
\begin{tabularx}{\textwidth}{XXX}
\toprule
\textbf{Question} & \textbf{Cluster 0 Response} & \textbf{Overall Population Response} \\
\midrule
\textbf{How much, if at all, do you think the following proposals would do to reduce economic inequality in the U.S.? Expanding government benefits for the poor} & A great deal: 0.0\% \newline A fair amount: 2.27\% \newline \textbf{Not too much: 71.55\%} \newline Nothing at all: 26.16\% & A great deal: 34.19\% \newline \textbf{A fair amount: 43.7\%} \newline Not too much: 19.83\% \newline Nothing at all: 2.26\% \\
\midrule
\textbf{How much, if at all, do you think the ease with which people can legally obtain guns contributes to gun violence in the country today?} & A great deal : 0.11\% \newline A fair amount: 22.86\% \newline \textbf{Not too much: 72.46\%} \newline Not at all: 4.55\% &  \textbf{A great deal : 46.86\%} \newline A fair amount: 38.02\% \newline Not too much: 14.51\% \newline Not at all: 0.6\% \\
\midrule
\textbf{How well does the Democratic party represent the interests of people like you?} & Very well: 0.0\% \newline Somewhat well: 0.0\% \newline Not too well: 11.94\% \newline \textbf{Not at all well: 88.05\%} & Very well: 4.97\% \newline \textbf{Somewhat well: 44.87\%} \newline Not too well: 31.93\% \newline Not at all well: 18.21\% \\
\bottomrule
\end{tabularx}
\label{tb:cls-pop-0}
\end{table*}

\begin{table*}[ht]
\small
\centering
\caption{Questions pertaining to immigration that exhibit the most disagreement among clusters}
\begin{tabularx}{\textwidth}{XXX}
\toprule
\textbf{Question} & \textbf{Cluster} & \textbf{Response} \\
\midrule
If you were deciding what the federal government should do to improve the quality of life for future generations, what priority would you give to who come here legally? Allowing more immigrants into the U.S. & Cluster-0 \newline Cluster-1 \newline Cluster-2 \newline Cluster-3 \newline Cluster-4 \newline Cluster-5 
& A lower priority (72.81\%)
\newline Top priority (99.05\%)
\newline Important, but not top priority (80.44\%)
\newline Important, but not top priority (89.87\%)
\newline Important, but not top priority (63.92\%)
\newline Top priority (69.51\%) \\
\midrule
How much, if at all, do you think the following proposals would do to reduce economic inequality in the U.S.? Reducing illegal immigration & Cluster-0 \newline Cluster-1 \newline Cluster-2 \newline Cluster-3 \newline Cluster-4 \newline Cluster-5 & A great deal (60.63\%) 
\newline A great deal (95.58\%) 
\newline Not too much (68.95\%) 
\newline Not too much (68.67\%) 
\newline A fair amount (69.41\%) 
\newline A fair amount (69.67\%) \\
\bottomrule
\end{tabularx}
\label{tb:immi}
\end{table*}

\textbf{Demographic Composition of Clusters:} We calculate the demographic statistics for each cluster and summarize the results for Political Party, Race and Education composition in Figure~\ref{fig:demo_dist} (compositions of other traits are summarized in Figure~\ref{fig:demo_dist_app} in the Appendix~\ref{sec:more_results}). Our analysis reveals that these clusters have distinct demographic compositions. For instance, in terms of political party, cluster-0 and cluster-1 lean toward Republican, while all other clusters lean toward Democrat. In terms of race, cluster-5 is dominated by Black and Hispanic races, while other clusters are dominated by the White race as in the overall population. In terms of education level, while cluster-0 and cluster-1 are dominated by members who received a college level education,  cluster 2 and cluster-3 have a majority of members that have received post-graduate level education. In cluster-4 and cluster-5, the majority of members only received a high school level education. 
We also find that clusters are composed of a mixture of different demographic groups, which corroborates the fact that individuals with different demographic traits can hold similar opinions over many topics. This further verifies the necessity of defining groups based on actual opinion as opposed to depending solely on demographic characteristics.

\textbf{Questions for which a cluster disagrees with the overall Population:} To further investigate the characteristics of each cluster, we show questions where a cluster mostly disagrees with the opinion of the overall population. For each cluster, we calculate the response distribution to each question, where we denote the distribution of cluster $c$ over question $q$ as $D_{c,q}$ for $c\in [5], q\in \mathcal{Q}$. Similarly, we calculate the overall population response distribution over each question and denote it as $D_{q}$. For each question $q \in \mathcal{Q}$, we calculate the Total Variation (TV) between the cluster response distribution and the overall population response distribution $TV(D_{i,q}, D_{q}), i \in [5]$.

Note that a larger value of $TV(q)$ indicates a greater degree of disagreement between the cluster and the overall population for a given question. In Table~\ref{tb:cls-pop-0}, we show the three questions with the largest total variation between cluster response distribution and overall population response distribution for cluster-0. We show the corresponding questions for clusters 1 to 5 in Tables~\ref{tb:cls-pop-1} to~\ref{tb:cls-pop-5} respectively, in the Appendix. Taking Cluster-0 as an example, we find that its members feel that expanding government benefits for the poor will not reduce economic inequality in the US, while we see over 75\% individuals in the overall population believe helping the poor helps to reduce the economic inequality a great deal or a fair amount. They also believe that allowing people to easily obtain guns legally, does not contribute to gun violence in the country. In contrast, over 46\% of the overall population believes that it does contribute to gun violence. Finally, when queried on how well the democratic party represented the interests of its members, 88.05\% of individuals in cluster-0 responded with {\it Not at all well}. In contrast, only 18.21\% of the overall population choose this option. Given that cluster-0 is mainly composed of Republicans, it is not surprising that the group has a clear mistrust of Democrats.

\textbf{Questions that differentiate clusters:} Finally, we show questions that elicit the most varied responses across the different clusters we have identified. More formally, we compute the average total variation between cluster pairs for each question as below:
\begin{align}
    TV_{Ave}(q) = \frac{1}{\binom{6}{2}}\sum_{0\leq i < j \leq 5} TV(D_{i,q}, D_{j,q})
\end{align}
Note that a larger value of $TV_{Ave}(q)$ indicates a greater degree of disagreement among clusters for that particular question.  In Table~\ref{tb:immi}, we present questions related to immigration topic that exhibit the greatest average total variation across clusters. The responses to these questions show interesting characteristics of each cluster. When considering the Immigration related topics, cluster-0, cluster-1 and cluster-5 have polarized attitudes, while the attitudes from other clusters are milder. In particular, Cluster-0 believes that a low priority should be assigned to allowing more legal immigrants, while placing importance on reducing the number of illegal immigrants. Cluster-1 and Cluster-5 believe that allowing more legal immigrants should be a top priority and that it is crucial to reduce illegal immigration. Cluster-2 and Cluster-3 assert that allowing more legal immigrants is not a top priority, and that reducing illegal immigration does not significantly affect economic inequality. Cluster-4 believes that allowing more legal immigrants is not a top priority. However, it also believes that reducing illegal immigrants could reduce economic inequality. In addition to questions related to immigration, the clusters also show disagreement in other questions belonging to topics such as the Crime (Table~\ref{tb:gun}) and Race (Table~\ref{tb:race}).

\begin{table*}[ht]
\small
\centering
\caption{Prediction accuracy across baselines and our experimental method.}
\begin{tabular}{lcccc}
\toprule
\textbf{Model} & \textbf{Raw Q.} & \textbf{Demographic} &\textbf{Context} &\textbf{Individual} \\
& & \textbf{+ Raw Q.} &\textbf{+ Raw Q.} &\textbf{Embeddings (Ours)} \\
\midrule
GPT-Neo-1.3B & 32.54\% & 33.40\% & 33.82\% & \textbf{59.99}\% \\
GPT-Neo-2.7B & 31.09\% & 34.32\%& 30.43\%& \textbf{60.59}\% \\
GPT-j-6B & 26.50\% & 31.86\%& 39.34\% & \textbf{61.84}\% \\
Falcon-7B-Instruct & 36.10\% & 38.40\%& 37.96\% & \textbf{60.78\%} \\
\bottomrule
\end{tabular}
\label{tb:res-comp}
\end{table*}

\begin{table}[ht]
\small
\centering
\caption{Prediction accuracy with and without training the SPM.}
\begin{tabular}{lcc}
\toprule
\textbf{Model} & \textbf{Random SPM}  &\textbf{Trained SPM} \\
& \textbf{Weights} & \textbf{Weights} \\
\midrule
GPT-Neo-1.3B & 7.92\% &\textbf{59.99}\% \\
GPT-Neo-2.7B & 8.31\% &\textbf{60.59}\% \\
GPT-j-6B & 23.26\%  &\textbf{61.84}\% \\
Falcon-7B-Instruct & 38.20\% &\textbf{60.78}\% \\
\bottomrule
\end{tabular}
\label{tb:abla-no-train}
\end{table}

\begin{table*}[ht]
\small
\centering
\caption{Prediction accuracy with different types of personas. For Demographic Embedding results we present only the political party trait (PARTY) which consists of 6 groups (see other demographic traits in Table~\ref{tb:abla-demo}).}
\begin{tabular}{lcccccc}
\toprule
\textbf{Model} & \textbf{Demographic Embeddings} &\textbf{Cluster Embeddings} &\textbf{Individual Embeddings} \\
& \textbf{[PARTY (6)]} & \textbf{[6 Clusters]} & \textbf{(Ours)}\\
\midrule
GPT-Neo-1.3B & 55.16\% & 55.57\% &\textbf{59.99}\% \\
GPT-Neo-2.7B & 55.66\% & 55.79\% &\textbf{60.59}\% \\
GPT-j-6B & 55.64\% & 55.82\% &\textbf{61.84}\% \\
Falcon-7B-Instruct & 52.87\% & 54.24\%&\textbf{60.78}\% \\
\bottomrule
\end{tabular}
\label{tb:abla-group-persona}
\end{table*}

\begin{table*}[ht]
\small
\centering
\caption{Prediction accuracy computed on unseen individuals for baselines and our method ($K$ is the number of responses we use to generate embeddings for unseen individuals).}
\begin{tabular}{lccccccccc}
\toprule
\textbf{Model} & \textbf{Raw Q.} & \textbf{Demographic} &\textbf{Context} & \textbf{K = 1} & \textbf{K = 5} &\textbf{K = 10} &\textbf{K = 20} &\textbf{K = 50} & \textbf{K = 100} \\
& & \textbf{+ Raw Q.} &\textbf{+ Raw Q.} \\
\midrule
GPT-Neo-1.3B & 32.22\% & 33.57\% & 33.97\% & 40.02\%  & 46.83\%  & 49.97\% & 53.27\%  & 56.18\% & 57.74\% \\
GPT-Neo-2.7B & 31.57\% & 33.94\% & 30.54\% & 40.40\%  & 46.58\% & 48.75\%  & 50.90\% & 52.66\% & 53.29\% \\
GPT-j-6B & 26.58\% & 32.10\% & 39.32\% & 41.41\%  & 49.29\% & 52.19\%  & 55.08\% & 57.73\% & 58.75\% \\
Falcon-7B-Instruct & 36.15\% & 38.22\% & 38.12\% & 39.11\%  & 48.12\%  & 51.16\%  & 53.87\% & 56.76\% & 57.83\% \\
\bottomrule
\end{tabular}
\label{tb:res-comp-strong}
\vspace{-0.2in}
\end{table*}

\subsection{Steering LLMs towards personas}
\label{sec:steer-exp-setting}
In this section, we benchmark the steerability of various LLMs toward personas defined in Sections~\ref{sec:CF}. We follow the procedure described in Section~\ref{sec:steer} and use prediction accuracy for evaluating performance. Prediction accuracy, in this case, is defined as the macro average of individual prediction accuracy (i.e., the percentage of questions where the LLM correctly predicts a particular individual's responses, averaged over all individuals). 

We compare the performance of our algorithm against the following baselines: 
\begin{enumerate}
\item \textbf{Raw Q.}~\cite{santurkar2023whose}: LLMs are only prompted with questions.
\item \textbf{Demographics + Raw Q.}~\cite{santurkar2023whose, hwang2023aligning}: the demographic traits of each individual are provided as context information.
\item \textbf{Context + Raw Q.}~\cite{hwang2023aligning}: a set of responses by the individual are provided as context information.
\end{enumerate}
We randomly split the individuals in the OpinionQA dataset into train ($\mathcal{P}_{tr}$) and evaluation ($\mathcal{P}_{val}$) partitions (represented in blue and green respectively, in Figure~\ref{fig:steer-structure}). Accordingly, the responses in each partition can be referred to as $\mathcal{R}_{tr} = \{(i,j), \text{where } r_{i,j} \text{ is not null}, i \in \mathcal{P}_{tr}, j \in \mathcal{Q}\}$ and $\mathcal{R}_{val} = \{(i,j), \text{where } r_{i,j} \text{ is not null}, i \in \mathcal{P}_{val}, j \in \mathcal{Q}\}$. Next, we randomly split the response sets in each partition above, into their own train and validation sets, which leaves us with the four sets that can be denoted as $\mathcal{R}_{tr}^{tr}$, $\mathcal{R}_{tr}^{val}$, $\mathcal{R}_{val}^{tr}$ and $\mathcal{R}_{val}^{val}$ (represented in light-blue, dark-blue, light-green and dark-green respectively, in Figure~\ref{fig:steer-structure}).

For our method, we first use the responses in $\mathcal{R}_{tr}^{tr}$ to optimize Eq.~\eqref{eq:CF} to get $U_{tr}^* = \{u_i \in \mathbb{R}^d, i\in\mathcal{P}_{tr}\}$ and $Q^* = \{q_j \in \mathbb{R}^d, j\in\mathcal{Q}\}$. Next, we use
$\{u_i \in \mathbb{R}^d, i\in\mathcal{P}_{tr}\}$ and $\mathcal{R}_{tr}^{tr}$ to train the SPM (Eq.~\eqref{eq:1}). We then report the average prediction accuracy of our steered LLMs over $\mathcal{R}_{tr}^{val}$  in Table~\ref{tb:res-comp}. For the \textbf{Context + Raw Q.} baseline, we use $\mathcal{R}_{tr}^{tr}$ to provide context questions. In particular, for a given question, we identify the $K$ most closely related questions within the training set, that the individual has responded to, to serve as the context. The similarity of these questions is measured using cosine distance bewteen the embeddings of the question (we employ the \texttt{text-embedding-ada-002}~\cite{openai_2023_new} model created by OpenAI to obtain the embeddings). The \textbf{Raw Q.} and \textbf{Demographics + Raw Q.} baselines do not use the train split. As in the case of our steered LLMs, we report the average prediction accuracy over $\mathcal{R}_{tr}^{val}$ for all baselines.

\subsubsection{Individual opinion prediction} 
In Table~\ref{tb:res-comp}, we present the average prediction accuracy of our method compared against the baselines. Note that for the \textbf{Demographics + Raw Q.} baseline, we include 13 different demographic traits in the prompt, and for the \textbf{Context + Raw Q.} baseline, we set $K = 5$ (an ablation study of different $K$ values is presented in Table~\ref{tb:abla-K}). Our method outperforms all baselines significantly, with improvements of $57\%-77\%$ over the best performing baseline for each model. We observe that when LLMs are provided with only the questions, their responses have low alignment with the individuals' responses. Including demographics traits in addition to responses to related questions tends to improve alignment. However, the LLMs still lack information about the individual which causes lower prediction accuracy. In contrast, our method, which steers the LLM utilizing the embeddings learned via CF, that encode knowledge about the overall opinions of a given individual, show higher prediction accuracy.

\subsubsection{Effectiveness of the SPM}
We verify the effectiveness of training the SPM by comparing the prediction accuracy with and without SPM training. As shown in Table~\ref{tb:abla-no-train}, the prediction accuracy clearly increases when the SPM is trained. This justifies the use of the SPM to map individual embeddings to the LLM embedding space.

\subsubsection{Performance of cluster personas}
We explore how well cluster persona embeddings can predict an individual's opinion. For this, we replace an individual's embedding with their corresponding cluster embedding (\textbf{CE}). For an individual belonging to cluster $i$, their corresponding cluster embedding is the centroid of the embeddings of {\it all} individuals belonging to the cluster $i$. The prediction accuracy is shown in the last two columns of Table~\ref{tb:abla-group-persona}. We see \textbf{CE}s get close performance as individual embeddings (around $8\%-12\%$ difference). Furthermore, we also compare \textbf{CE}s with the demographic group embedding (\textbf{DE}) in Table~\ref{tb:abla-group-persona}. \textbf{DE} is the centroid of individual embeddings belonging to the same demographic group. For instance, for an individual who belongs to the Democrat party, we use a demographic embedding that is the average of the embeddings of all the individuals who belong to the Democrat party. We see that the use of \textbf{CE}s provide improvements of between $0.23\%-2.59\%$ over \textbf{DE}s. Note that for \textbf{DE}s, we consider the political party trait that consists of six parties and shows the highest prediction accuracy, compared to other demographic traits (shown separately in Table~\ref{tb:abla-demo} in the Appendix). For \textbf{CE}s, we present results using six clusters in Table~\ref{tb:abla-group-persona}, which enables us to obtain better prediction accuracy than any demographic trait. As shown in Table~\ref{tb:abla-cls} in the Appendix, increasing the number of clusters can yield much higher prediction accuracies. 



\subsubsection{Generalization to unseen individuals} 
We also run an ablation test to see if the steered LLM can generalize to \textbf{unseen individuals}. Here, we use responses from the set $\mathcal{R}_{val}^{tr}$ to get embeddings $U_{val}^*$ for individuals in $\mathcal{P}_{val}$. $\mathcal{R}_{val}^{tr}$ contains a small number of responses for each individual and the question embeddings are fixed as $Q^*$ (learned based on $\mathcal{P}_{tr}$ as defined in Section~\ref{sec:steer-exp-setting}). Note that the SPM, trained on $\mathcal{P}_{tr}$, has not seen any responses from $\mathcal{P}_{val}$. We explore the use of different numbers of responses ($K$) from $\mathcal{R}_{val}^{tr}$ to generate the embeddings for the unseen individuals. In Table~\ref{tb:res-comp-strong},
we report the prediction accuracies for our models over $\mathcal{R}_{val}^{val}$ for these experiments and for our baselines. We see, even in the case of using a single response per individual, our method generalizes better to unseen individuals compared to the baselines. Furthermore, the prediction accuracies increase as we use more responses ($K$) to get more accurate embeddings. See Appendix~\ref{sec:more_results} for further results.

\section{Conclusion}
In this work, we present an approach for steerable generation using LLMs, where we utilize a data-driven notion of a persona to modulate the generation process. We proposed an simple model and efficient algorithm to align LLMs with these personas (both individuals and clusters). We validate the efficacy of our algorithm using the OpinionQA dataset. For a select set of LLMs we show that our method out-performs traditional steering mechanisms supporting our hypothesis that LLMs align with individuals' opinions better when leveraging our data-driven personas.

\section*{Limitations}
There are some limitations to the work we present here. Firstly, we rely on the QA format to perform collaborative filtering and embed individuals into an embedding space. Secondly, we only test our approach over one dataset due to time and resource constraints. However, we note that our method should scale well to other similar datasets. Thirdly, in this work we only test prefx-tuning \citep{lester2021power} and prompt-tuning \citep{lester2021power} for steering LLMs. Other parameter efficient fine tuning methods such as LoRA \citep{hu2021lora}, (IA)$^3$~\cite{liu2022few} \emph{etc.} can also be incorporated into our approach.
 
\section*{Ethical Considerations}
We fine tune the LLMs over a dataset encoding the opinions of individuals and we acknowledge that the LLMs could reflect and even intensify the biases held by certain individuals in the training set. A careful audit to the training data is necessary before training in practice.

\begin{figure*}[ht]
    \centering
    \includegraphics[width=0.33\linewidth]{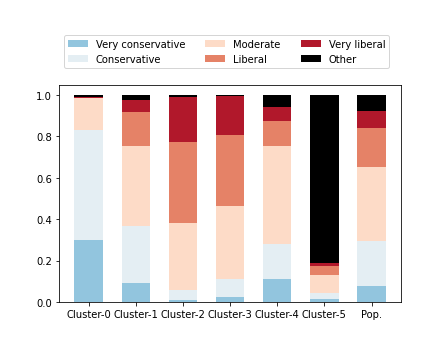}
    \includegraphics[width=0.33\linewidth]{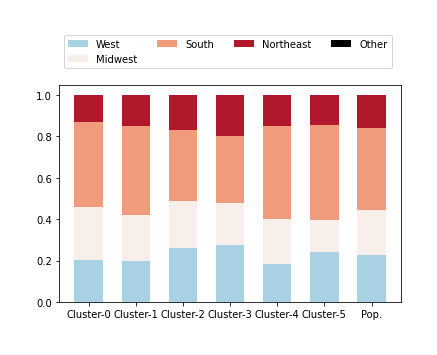}
    \includegraphics[width=0.33\linewidth]{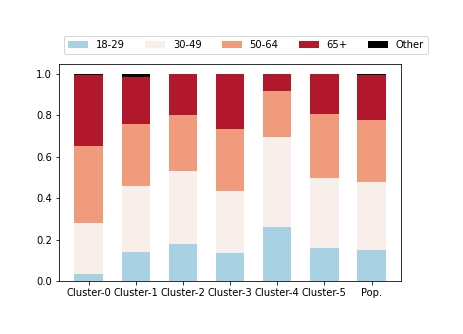}
    \includegraphics[width=0.33\linewidth]{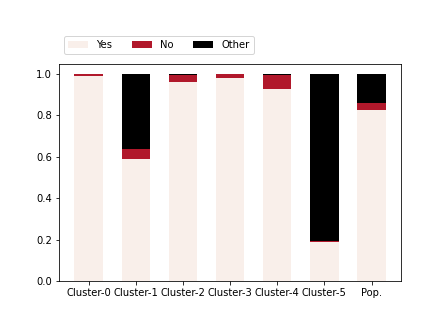}
    \includegraphics[width=0.33\linewidth]{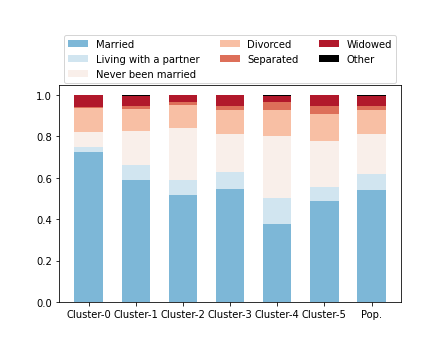}
    \includegraphics[width=0.33\linewidth]{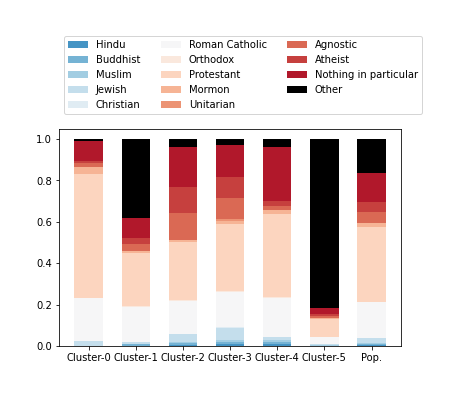}
    \includegraphics[width=0.33\linewidth]{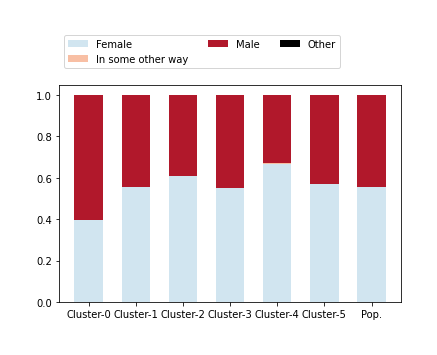}
    \includegraphics[width=0.33\linewidth]{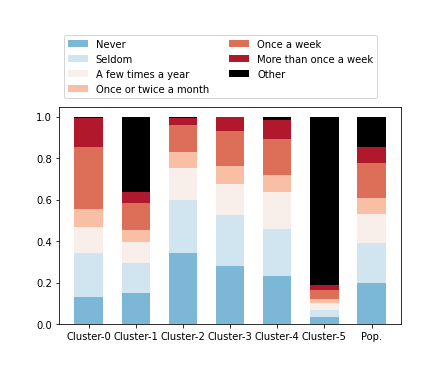}
    \includegraphics[width=0.33\linewidth]{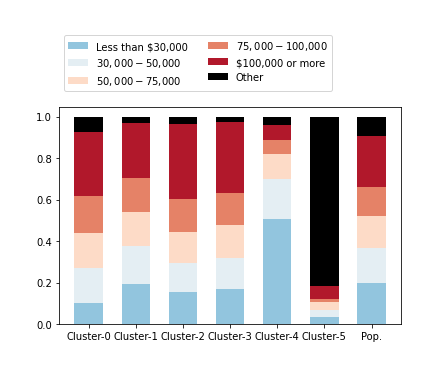}
    
    \caption{The demographic composition of Clusters-0 to 5 and the Overall Population. From left to right and top to bottom, we show demographic composition for Ideology, Region, Age, Citizenship, Marital status, Religion, Sex, Religion attendance and Income.}
    \label{fig:demo_dist_app}
\end{figure*}

\bibliography{anthology,custom}

\begin{thebibliography}{30}
\expandafter\ifx\csname natexlab\endcsname\relax\def\natexlab#1{#1}\fi

\bibitem[{Argyle et~al.(2023)Argyle, Busby, Fulda, Gubler, Rytting, and Wingate}]{argyle2023out}
Lisa~P Argyle, Ethan~C Busby, Nancy Fulda, Joshua~R Gubler, Christopher Rytting, and David Wingate. 2023.
\newblock Out of one, many: Using language models to simulate human samples.
\newblock \emph{Political Analysis}, 31(3):337--351.

\bibitem[{Bishop(2006)}]{bishop2006pattern}
Christopher~M. Bishop. 2006.
\newblock \emph{Pattern Recognition and Machine Learning}.
\newblock Springer.

\bibitem[{Brown et~al.(2020)Brown, Mann, Ryder, Subbiah, Kaplan, Dhariwal, Neelakantan, Shyam, Sastry, Askell et~al.}]{brown2020language}
Tom Brown, Benjamin Mann, Nick Ryder, Melanie Subbiah, Jared~D Kaplan, Prafulla Dhariwal, Arvind Neelakantan, Pranav Shyam, Girish Sastry, Amanda Askell, et~al. 2020.
\newblock Language models are few-shot learners.
\newblock \emph{Advances in neural information processing systems}, 33:1877--1901.

\bibitem[{Deshpande et~al.(2023)Deshpande, Murahari, Rajpurohit, Kalyan, and Narasimhan}]{deshpande2023toxicity}
Ameet Deshpande, Vishvak Murahari, Tanmay Rajpurohit, Ashwin Kalyan, and Karthik Narasimhan. 2023.
\newblock Toxicity in chatgpt: Analyzing persona-assigned language models.
\newblock \emph{arXiv preprint arXiv:2304.05335}.

\bibitem[{Durmus et~al.(2023)Durmus, Nyugen, Liao, Schiefer, Askell, Bakhtin, Chen, Hatfield-Dodds, Hernandez, Joseph et~al.}]{durmus2023towards}
Esin Durmus, Karina Nyugen, Thomas~I Liao, Nicholas Schiefer, Amanda Askell, Anton Bakhtin, Carol Chen, Zac Hatfield-Dodds, Danny Hernandez, Nicholas Joseph, et~al. 2023.
\newblock Towards measuring the representation of subjective global opinions in language models.
\newblock \emph{arXiv preprint arXiv:2306.16388}.

\bibitem[{Feng et~al.(2023)Feng, Park, Liu, and Tsvetkov}]{feng2023pretraining}
Shangbin Feng, Chan~Young Park, Yuhan Liu, and Yulia Tsvetkov. 2023.
\newblock From pretraining data to language models to downstream tasks: Tracking the trails of political biases leading to unfair nlp models.
\newblock \emph{arXiv preprint arXiv:2305.08283}.

\bibitem[{Hu et~al.(2021)Hu, Shen, Wallis, Allen-Zhu, Li, Wang, Wang, and Chen}]{hu2021lora}
Edward~J Hu, Yelong Shen, Phillip Wallis, Zeyuan Allen-Zhu, Yuanzhi Li, Shean Wang, Lu~Wang, and Weizhu Chen. 2021.
\newblock Lora: Low-rank adaptation of large language models.
\newblock \emph{arXiv preprint arXiv:2106.09685}.

\bibitem[{Hwang et~al.(2023)Hwang, Majumder, and Tandon}]{hwang2023aligning}
EunJeong Hwang, Bodhisattwa~Prasad Majumder, and Niket Tandon. 2023.
\newblock Aligning language models to user opinions.
\newblock \emph{arXiv preprint arXiv:2305.14929}.

\bibitem[{Jiang et~al.(2022)Jiang, Beeferman, Roy, and Roy}]{jiang2022communitylm}
Hang Jiang, Doug Beeferman, Brandon Roy, and Deb Roy. 2022.
\newblock Communitylm: Probing partisan worldviews from language models.
\newblock \emph{arXiv preprint arXiv:2209.07065}.

\bibitem[{Lester et~al.(2021)Lester, Al-Rfou, and Constant}]{lester2021power}
Brian Lester, Rami Al-Rfou, and Noah Constant. 2021.
\newblock The power of scale for parameter-efficient prompt tuning.
\newblock \emph{arXiv preprint arXiv:2104.08691}.

\bibitem[{Li and Liang(2021)}]{li2021prefix}
Xiang~Lisa Li and Percy Liang. 2021.
\newblock Prefix-tuning: Optimizing continuous prompts for generation.
\newblock \emph{arXiv preprint arXiv:2101.00190}.

\bibitem[{Liu et~al.(2022)Liu, Tam, Muqeeth, Mohta, Huang, Bansal, and Raffel}]{liu2022few}
Haokun Liu, Derek Tam, Mohammed Muqeeth, Jay Mohta, Tenghao Huang, Mohit Bansal, and Colin~A Raffel. 2022.
\newblock Few-shot parameter-efficient fine-tuning is better and cheaper than in-context learning.
\newblock \emph{Advances in Neural Information Processing Systems}, 35:1950--1965.

\bibitem[{Liu et~al.(2021)Liu, Zheng, Du, Ding, Qian, Yang, and Tang}]{liu2021gpt}
Xiao Liu, Yanan Zheng, Zhengxiao Du, Ming Ding, Yujie Qian, Zhilin Yang, and Jie Tang. 2021.
\newblock Gpt understands, too.
\newblock \emph{arXiv preprint arXiv:2103.10385}.

\bibitem[{Loshchilov and Hutter(2017)}]{loshchilov2017decoupled}
Ilya Loshchilov and Frank Hutter. 2017.
\newblock Decoupled weight decay regularization.
\newblock \emph{arXiv preprint arXiv:1711.05101}.

\bibitem[{{OpenAI}(2023)}]{openai_2023_new}
{OpenAI}. 2023.
\newblock New and improved embedding model.
\newblock \url{https://openai.com/blog/new-and-improved-embedding-model}.
\newblock Accessed: [insert date of access].

\bibitem[{OpenAI(2023)}]{openai2023gpt}
R~OpenAI. 2023.
\newblock Gpt-4 technical report.
\newblock \emph{arXiv}, pages 2303--08774.

\bibitem[{Ouyang et~al.(2022)Ouyang, Wu, Jiang, Almeida, Wainwright, Mishkin, Zhang, Agarwal, Slama, Ray et~al.}]{ouyang2022training}
Long Ouyang, Jeffrey Wu, Xu~Jiang, Diogo Almeida, Carroll Wainwright, Pamela Mishkin, Chong Zhang, Sandhini Agarwal, Katarina Slama, Alex Ray, et~al. 2022.
\newblock Training language models to follow instructions with human feedback.
\newblock \emph{Advances in Neural Information Processing Systems}, 35:27730--27744.

\bibitem[{Ozdayi et~al.(2023)Ozdayi, Peris, FitzGerald, Dupuy, Majmudar, Khan, Parikh, and Gupta}]{Ozdayi2023ControllingTE}
Mustafa~Safa Ozdayi, Charith~S. Peris, Jack G.~M. FitzGerald, Christophe Dupuy, Jimit Majmudar, Haidar Khan, Rahil Parikh, and Rahul Gupta. 2023.
\newblock \href {https://api.semanticscholar.org/CorpusID:258823013} {Controlling the extraction of memorized data from large language models via prompt-tuning}.
\newblock In \emph{Annual Meeting of the Association for Computational Linguistics}.

\bibitem[{Salewski et~al.(2023)Salewski, Alaniz, Rio-Torto, Schulz, and Akata}]{salewski2023context}
Leonard Salewski, Stephan Alaniz, Isabel Rio-Torto, Eric Schulz, and Zeynep Akata. 2023.
\newblock In-context impersonation reveals large language models' strengths and biases.
\newblock \emph{arXiv preprint arXiv:2305.14930}.

\bibitem[{Santurkar et~al.(2023)Santurkar, Durmus, Ladhak, Lee, Liang, and Hashimoto}]{santurkar2023whose}
Shibani Santurkar, Esin Durmus, Faisal Ladhak, Cinoo Lee, Percy Liang, and Tatsunori Hashimoto. 2023.
\newblock Whose opinions do language models reflect?
\newblock \emph{arXiv preprint arXiv:2303.17548}.

\bibitem[{Santy et~al.(2023)Santy, Liang, Bras, Reinecke, and Sap}]{santy2023nlpositionality}
Sebastin Santy, Jenny~T Liang, Ronan~Le Bras, Katharina Reinecke, and Maarten Sap. 2023.
\newblock Nlpositionality: Characterizing design biases of datasets and models.
\newblock \emph{arXiv preprint arXiv:2306.01943}.

\bibitem[{Scherrer et~al.(2023)Scherrer, Shi, Feder, and Blei}]{scherrer2023evaluating}
Nino Scherrer, Claudia Shi, Amir Feder, and David~M Blei. 2023.
\newblock Evaluating the moral beliefs encoded in llms.
\newblock \emph{arXiv preprint arXiv:2307.14324}.

\bibitem[{Shazeer(2019)}]{shazeer2019fast}
Noam Shazeer. 2019.
\newblock Fast transformer decoding: One write-head is all you need.
\newblock \emph{arXiv preprint arXiv:1911.02150}.

\bibitem[{Simmons(2022)}]{simmons2022moral}
Gabriel Simmons. 2022.
\newblock Moral mimicry: Large language models produce moral rationalizations tailored to political identity.
\newblock \emph{arXiv preprint arXiv:2209.12106}.

\bibitem[{Singhal et~al.(2023)Singhal, Azizi, Tu, Mahdavi, Wei, Chung, Scales, Tanwani, Cole-Lewis, Pfohl et~al.}]{singhal2023large}
Karan Singhal, Shekoofeh Azizi, Tao Tu, S~Sara Mahdavi, Jason Wei, Hyung~Won Chung, Nathan Scales, Ajay Tanwani, Heather Cole-Lewis, Stephen Pfohl, et~al. 2023.
\newblock Large language models encode clinical knowledge.
\newblock \emph{Nature}, pages 1--9.

\bibitem[{Tan et~al.(2023)Tan, Pang, and Fan}]{tan2023towards}
Kehui Tan, Tianqi Pang, and Chenyou Fan. 2023.
\newblock Towards applying powerful large ai models in classroom teaching: Opportunities, challenges and prospects.
\newblock \emph{arXiv preprint arXiv:2305.03433}.

\bibitem[{Touvron et~al.(2023)Touvron, Martin, Stone, Albert, Almahairi, Babaei, Bashlykov, Batra, Bhargava, Bhosale et~al.}]{touvron2023llama}
Hugo Touvron, Louis Martin, Kevin Stone, Peter Albert, Amjad Almahairi, Yasmine Babaei, Nikolay Bashlykov, Soumya Batra, Prajjwal Bhargava, Shruti Bhosale, et~al. 2023.
\newblock Llama 2: Open foundation and fine-tuned chat models.
\newblock \emph{arXiv preprint arXiv:2307.09288}.

\bibitem[{Wu et~al.(2023)Wu, Irsoy, Lu, Dabravolski, Dredze, Gehrmann, Kambadur, Rosenberg, and Mann}]{wu2023bloomberggpt}
Shijie Wu, Ozan Irsoy, Steven Lu, Vadim Dabravolski, Mark Dredze, Sebastian Gehrmann, Prabhanjan Kambadur, David Rosenberg, and Gideon Mann. 2023.
\newblock Bloomberggpt: A large language model for finance.
\newblock \emph{arXiv preprint arXiv:2303.17564}.

\bibitem[{Zhang et~al.(2023)Zhang, Han, Zhou, Hu, Yan, Lu, Li, Gao, and Qiao}]{zhang2023llama}
Renrui Zhang, Jiaming Han, Aojun Zhou, Xiangfei Hu, Shilin Yan, Pan Lu, Hongsheng Li, Peng Gao, and Yu~Qiao. 2023.
\newblock Llama-adapter: Efficient fine-tuning of language models with zero-init attention.
\newblock \emph{arXiv preprint arXiv:2303.16199}.

\bibitem[{Zhao et~al.(2023)Zhao, Zhou, Li, Tang, Wang, Hou, Min, Zhang, Zhang, Dong et~al.}]{zhao2023survey}
Wayne~Xin Zhao, Kun Zhou, Junyi Li, Tianyi Tang, Xiaolei Wang, Yupeng Hou, Yingqian Min, Beichen Zhang, Junjie Zhang, Zican Dong, et~al. 2023.
\newblock A survey of large language models.
\newblock \emph{arXiv preprint arXiv:2303.18223}.

\end{thebibliography}

\newpage
\appendix

\section{Additional details on the Dataset and the Experiments}
\subsection{Additional information on the OpinionQA dataset}\label{sec:more_data}
We consider the coarse topics of OpinionQA which include: healthcare system,
future, relationships and family, leadership,
status in life, political issues, personal health, race,
personal finance, crime or security, corporations, banks, technology and automation, self-perception and values,
science, education, religion, discrimination,
community health, immigration, global attitudes and foreign policy, economy and inequality, news, social media, data, privacy, job/career, gender and sexuality. The list of demographic traits available for each participant are: Education, Citizen, Marital, Income, Political Ideology, Region, Political Party, Sex, Age, Religion Attendance, Race and Religion.

\subsection{Details of optimizing Eq.~\eqref{eq:CF}}
To optimize Eq.~\eqref{eq:CF}, we perform mini-batch gradient descent with the Adam optimizer (learning rate 0.001) and a batch-size of 2048. Furthermore, both individuals and questions are represented by 16-dimensional vector. For unseen users, we fix the question embedding and only optimize the individual embedding for new users.

\subsection{Architecture of the SPM}
The SPM that we use is a two-layer multilayer perceptron (MLP) with 32 hidden units. The output of the SPM is a vector of shape [$T$, $L$$\times$2$\times$$D$] where T is the number of virtual tokens, L is the number of layers and D is the token dimension. We follow the setting of prefix-tuning~\cite{li2021prefix} to insert virtual tokens before each transformer layer. Since the Falcon model uses multi-query attention~\cite{shazeer2019fast}, the output of the SPM becomes a vector of shape [$T$, $L$$\times$2$\times$$D$/$H$] where H is the number of heads. We set the number of virtual tokens to be 1 in our experiments. For training, we use the  AdamW~\cite{loshchilov2017decoupled} optimizer with learning rate 0.001 and weight decay 0.001, and we train the SPM for 10 epochs with early stopping.

\subsection{Prompt templates of baseline methods}
\begin{figure*}[t]
    \centering
    \includegraphics[width=0.95\linewidth]{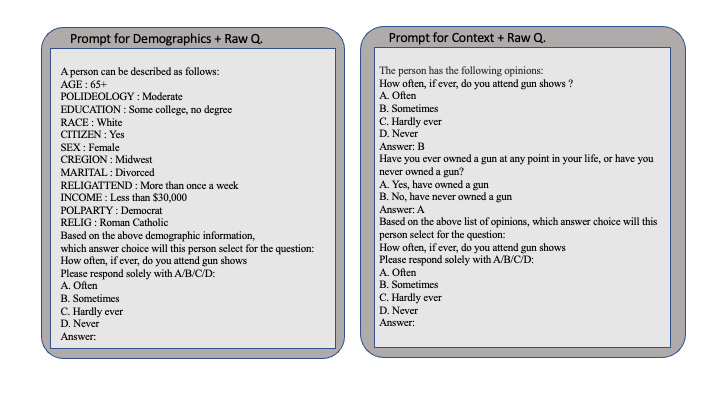}
    \caption{Example prompts of baseline methods. The left panel shows an example prompt used for the \textbf{Demographics + Raw Q.} baseline while the right panel shows an example prompt used for the \textbf{Context + Raw Q.} baseline.}
    \label{fig:prompt}
\end{figure*}

In Table~\ref{tb:res-comp}, we includes three types of baselines \textbf{Raw Q.}, \textbf{Demographics + Raw Q.} and \textbf{Context + Raw Q.}. For the \textbf{Raw Q.} baseline, only the question text is included in the prompt. For the \textbf{Demographics + Raw Q.} baseline, we prepend the demographics of an individual before the question text. An example prompt is shown on the left panel of Figure~\ref{fig:prompt}. For the \textbf{Context + Raw Q.} baseline, we prepend a set of responses provided by individual to other questions. An example prompt is shown on the right panel of Figure~\ref{fig:prompt}.

\subsection{Additional experimental results}
\label{sec:more_results}
We provide further detail on our experimental results in this section. 

In Figure~\ref{fig:demo_dist_app}, we show the demographic composition of clusters for other demographic traits not included in Figure~\ref{fig:demo_dist}. 

In Table~\ref{tb:abla-demo}, we show the prediction accuracy when we use a single demographic-trait as the representation of an individual. We note that our method with the use of individual or cluster embeddings out-performs all demographic embeddings. 

In Table~\ref{tb:abla-demo-comb}, we present results obtained by using combinations of demographic groups to represent individuals. We note that individual or cluster embeddings out-performs the prediction accuracy of these combinations as well.

In Table~\ref{tb:abla-K} we show the prediction accuracy for the \textbf{Context + Raw Q.} baseline under different $K$ values, \emph{i.e.} the number of context samples. 

As mentioned in Section~\ref{sec:steer}, we draw from {\it prefix-tuning} tuning technique to implement our SPM. In order test the robustness of our method to other prompting methods, we also implement the SPM with the use of prompt-tuning \citep{lester2021power}. We show our results in Table~\ref{tb:res-comp-full}.

Table~\ref{tb:cls-pop-1}-Table~\ref{tb:cls-pop-5} show the top-3 questions that a cluster mostly disagrees with, when considering the overall population. Finally, Table~\ref{tb:gun} and Table~\ref{tb:race} show the questions that have the largest disagreement among clusters for the Crime or Security and Immigration topics.


\begin{table*}[ht]
\small
\centering
\caption{Prediction accuracy across baselines and our experimental method showcasing the use of both {\it prefix-tuning} and {\it prompt-tuning} techniques.}
\begin{tabular}{lccccc} 
\toprule
\textbf{Model} & \textbf{Raw Q.} & \textbf{Demographic} &\textbf{Context} &\textbf{Prefix-Tuning} & \textbf{Prompt-Tuning} \\
& & \textbf{+ Raw Q.} &\textbf{+ Raw Q.} &\textbf{(Ours)} & \textbf{(Ours)} \\
\midrule
GPT-Neo-1.3B & 32.54\% & 33.40\% & 33.82\% & \textbf{59.99}\% & \textbf{59.65}\% \\
GPT-Neo-2.7B & 31.09\% & 34.32\%& 30.43\%& \textbf{60.59}\% & \textbf{58.96\%} \\
GPT-j-6B & 26.50\% & 31.86\%& 39.34\% & \textbf{61.84}\% & \textbf{59.41\%}\\
Falcon-7B-Instruct & 36.10\% & 38.40\%& 37.96\% & \textbf{60.78\%} & \textbf{57.98\%} \\
\bottomrule
\end{tabular}
\label{tb:res-comp-full}
\end{table*}

\begin{table*}[ht]
\small
    \centering
    \caption{Prediction accuracy for the Context + Raw Q baseline using different numbers of context samples \( K \).}
    \begin{tabular}{lcccc}
        \toprule
        \textbf{Model} & \( \textbf{K=3} \) & \( \textbf{K=5} \) & \( \textbf{K=8} \) & \( \textbf{K=10} \) \\
        \midrule
        GPT-Neo-1.3B & 33.05\% & 33.82\% & 31.43\% & 31.41\% \\
        GPT-Neo-2.7B & 32.99\% & 30.43\% & 29.76\% & 29.89\% \\
        GPT-j-6B & 37.88\% & 39.34\% & 38.32\% & 38.49\% \\
        Falcon-7B-Instruct & 36.78\%  &37.96\%  & 36.59\%  & 36.98\% \\
        \bottomrule
    \end{tabular}
\label{tb:abla-K}
\end{table*}

\begin{table*}[ht]
\small
\centering
\caption{Prediction accuracy across different types of demographic traits (Single). We provide the number of groups in each demographic trait within parenthesis}
\begin{tabular}{lccccccc}
\toprule
\textbf{Model} & \textbf{EDUCATION} & \textbf{INCOME} & \textbf{RACE} & \textbf{RELIGION} & \textbf{PARTY} & \textbf{Cluster} &\textbf{Individual} \\
& \textbf{(7)} & \textbf{(7)} & \textbf{(7)} & \textbf{(16)} & \textbf{(6)} & \textbf{Embeddings} &\textbf{Embeddings} \\
\midrule
GPT-Neo-1.3B & 53.24\% & 53.09\% & 53.02\% & 53.85\% & 55.16\% & 55.57\% & 59.99\% \\
GPT-Neo-2.7B & 53.69\% & 53.43\% & 53.35\% & 54.31\% & 55.66\% &55.79\% & 60.59\% \\
GPT-j-6B & 53.73\% & 53.25\% & 53.25\% & 54.32\% & 55.64\% & 55.82\% & 61.84\% \\
Falcon-7B-Instruct & 50.87\%  & 51.65\% & 51.14\%  & 51.78\% & 52.54\% & 54.24\% & 60.78\% \\
\bottomrule
\end{tabular}
\label{tb:abla-demo}
\end{table*}

\begin{table*}[ht]
\small
\centering
\caption{Prediction accuracy for different combinations of demographic traits.}
\begin{tabular}{lcccccc}
\toprule
\textbf{Model} & \textbf{EDUCATION X} & \textbf{INCOME X} &  \textbf{RELIGION X} & \textbf{Cluster} & \textbf{Individual} \\
\textbf{Model} & \textbf{PARTY (42)} & \textbf{PARTY (42)} &  \textbf{PARTY (96)} & \textbf{Embeddings} & \textbf{Embeddings} \\
\midrule
GPT-Neo-1.3B & 55.59\% & 55.78\% & 55.88\% & 55.57\% & 59.99\% \\
GPT-Neo-2.7B & 56.03\% & 56.06\% &  56.35\% & 55.79\% & 60.59\% \\
GPT-j-6B & 56.16\% & 56.14\% &  56.47\% & 55.82\% & 61.84\% \\
Falcon-7B-Instruct & 54.96\% & 54.78\% & 55.34\% & 54.24\% & 60.78\%  \\
\bottomrule
\end{tabular}
\label{tb:abla-demo-comb}
\end{table*}

\begin{table*}[ht]
\small
\centering
\caption{Prediction accuracy of cluster personas across different choices for number of clusters.}
\begin{tabular}{lcccccc}
\toprule
\textbf{Model} & \textbf{6 Clusters} & \textbf{10 Clusters} & \textbf{20 Clusters} & \textbf{30 Clusters} & \textbf{50 Clusters} & \textbf{Individual Embeddings} \\
\midrule
GPT-Neo-1.3B & 55.57\% & 56.65\% & 58.11\% & 58.41\% & 58.74\% & 59.99\% \\
GPT-Neo-2.7B & 55.79\% & 56.93\% & 58.61\% & 58.93\% & 59.42\% & 60.59\% \\
GPT-j-6B & 55.82\% & 57.13\% & 58.97\% & 59.52\% & 59.86\% & 61.84\% \\
Falcon-7B-Instruct &54.24\%  &56.72\%  &58.99\% & 59.18\%  & 59.39\%  & 60.78\% \\
\bottomrule
\end{tabular}
\label{tb:abla-cls}
\end{table*}

\begin{table*}[ht]
\small
\centering
\caption{Comparison of responses between Cluster 1 and the Overall Population.}
\begin{tabularx}{\textwidth}{XXX}
\toprule
\textbf{Question} & \textbf{Cluster 1 Response} & \textbf{Overall Population Response} \\
\midrule
\textbf{How much control, if any, do you think you have over who can access the search terms you use online?} & \textbf{A lot of control: 99.47\%} \newline Offline: does not have internet or A little control: 0.52\% \newline No control: 0.0\% & A lot of control: 25.3\% \newline \textbf{Offline: does not have internet or A little control: 60.88\%} \newline No control: 13.8\% \\
\midrule
\textbf{For each, please indicate if you, personally, think it is acceptable. A white person using the n-word} & \textbf{Always acceptable: 87.38\%} \newline Sometimes acceptable: 12.52\% \newline Rarely acceptable: 0.09\% \newline Never acceptable: 0.0\% & Always acceptable: 20.09\% \newline Sometimes acceptable: 2.95\% \newline Rarely acceptable: 25.03\% \newline \textbf{Never acceptable: 51.91\%} \\
\midrule
\textbf{Overall, how does being Jewish affect people's ability to get ahead in our country these days?} & \textbf{Helps a lot: 99.59\%} \newline Helps a little: 0.4\% \newline Neither helps nor hurts: 0.0\% \newline Hurts a little: 0.0\% \newline Hurts a lot: 0.0\% & Helps a lot: 22.9\% \newline Helps a little: 21.6\% \newline \textbf{Neither helps nor hurts: 52.89\%} \newline Hurts a little: 2.59\% \newline Hurts a lot: 0.0\% \\
\bottomrule
\end{tabularx}
\label{tb:cls-pop-1}
\end{table*}

\begin{table*}[ht]
\small
\centering
\caption{Comparison of responses between Cluster 2 and the Overall Population.}
\begin{tabularx}{\textwidth}{XXX}
\toprule
\textbf{Question} & \textbf{Cluster 2 Response} & \textbf{Overall Population Response} \\
\midrule
\textbf{How confident are you, if at all, that the actions taken by the international community will significantly reduce the effects of global climate change?} & Very confident: 1.18\% \newline \textbf{Somewhat confident: 91.61\%} \newline Not too confident: 7.19\% \newline Not at all confident: 0.0\% & Very confident: 25.32\% \newline \textbf{Somewhat confident: 41.1\%} \newline Not too confident: 31.73\% \newline Not at all confident: 1.83\% \\
\midrule
\textbf{In general, how often, if ever, would you say you have parties or get-togethers with any of your neighbors?} & Every day: 1.65\% \newline  Several times a week: 16.71\% \newline \textbf{About once a week: 49.86\%}  \newline About once a month: 30.25\% \newline Less than once a month: 1.51\% \newline Never: 0.0\% & Every day: 12.33\% \newline Several times a week: 11.1\% \newline About once a week: 17.87\% \newline  About once a month: 17.94\% \newline \textbf{Less than once a month: 28.73\%} \newline Never: 12.0\% \\
\midrule
\textbf{How enthusiastic are you, if at all, about the possibility of using computer programs to make hiring decisions for society as a whole?} & Very enthusiastic: 6.73\% \newline \textbf{Somewhat enthusiastic: 81.57\%} \newline Not too enthusiastic: 11.69\% \newline Not at all enthusiastic: 0.0\% & Very enthusiastic: 27.55\% \newline \textbf{Somewhat enthusiastic: 33.07\%} \newline Not too enthusiastic: 38.58\% \newline Not at all enthusiastic: 0.78\% \\
\bottomrule
\end{tabularx}
\label{tb:cls-pop-2}
\end{table*}

\begin{table*}[ht]
\small
\centering
\caption{Comparison of Responses between Cluster 3 and the Overall Population.}
\begin{tabularx}{\textwidth}{XXX}
\toprule
\textbf{Question} & \textbf{Cluster 3 Response} & \textbf{Overall Population Response} \\
\midrule
\textbf{How often, if ever, do you attend gun shows?} & \textbf{Never: 90.7\%} \newline Hardly ever: 9.29\% \newline Sometimes: 0.0\% \newline Often: 0.0\% & Never: 20.6\% \newline \textbf{Hardly ever: 39.74\%} \newline Sometimes: 20.36\% \newline Often: 19.29\% \\
\midrule
\textbf{How important, if at all, is being a gun owner to your overall identity?} & Not at all important: 5.14\% \newline \textbf{Not too important: 89.26\%} \newline Somewhat important: 5.58\% \newline  Very important: 0.0\% & Not at all important: 0.57\% \newline Not too important: 24.21\% \newline \textbf{Somewhat important: 42.63\%} \newline Very important: 32.57\% \\
\midrule
\textbf{Thinking about the country today, would you say there are?} & \textbf{Too few women in top executive business positions: 88.1\%} \newline About the right number of women in top executive business positions: 11.89\% \newline Too many women in top executive business positions: 0.0\% & Too few women in top executive business positions: 18.73\% \newline \textbf{About the right number of women in top executive business positions: 41.96\%} \newline Too many women in top executive business positions: 39.3\% \\
\bottomrule
\end{tabularx}
\label{tb:cls-pop-3}
\end{table*}

\begin{table*}[ht]
\small
\centering
\caption{Comparison of Responses between Cluster 4 and the Overall Population.}
\begin{tabularx}{\textwidth}{XXX}
\toprule
\textbf{Question} & \textbf{Cluster 4 Response} & \textbf{Overall Population Response} \\
\midrule
\textbf{Thinking about medical doctors, how often would you say they do a good job providing diagnoses and treatment recommendations?} & All or most of the time: 20.0\% \newline \textbf{Some of the time: 77.84\%} \newline Only a little of the time: 2.15\% \newline None of the time: 0.0\% & \textbf{All or most of the time: 70.3\%} \newline Some of the time: 29.61\% \newline Only a little of the time: 0.07\% \newline None of the time: 0.0\% \\
\midrule
\textbf{If robots and computers do much of the work currently done by humans, do you think this would be} & A very good thing for the country: 0.0\% \newline A somewhat good thing for the country: 15.88\% \newline  \textbf{A somewhat bad thing for the country : 76.66\%} \newline A very bad thing for the country: 7.45\% & A very good thing for the country: 23.29\% \newline A somewhat good thing for the country: 28.35\% \newline  \textbf{A somewhat bad thing for the country : 47.21\%} \newline A very bad thing for the country: 1.13\% \\
\midrule
\textbf{In the future, what kind of an impact do you think science and technology will have in solving the biggest problems facing the country?} & A very positive impact: 15.49\% \newline \textbf{A somewhat positive impact: 81.37\%} \newline A somewhat negative impact: 3.13\% \newline A very negative impact: 0.0\% & \textbf{A very positive impact: 63.59\%} \newline A somewhat positive impact: 36.1\% \newline A somewhat negative impact: 0.29\% \newline A very negative impact: 0.0\% \\
\bottomrule
\end{tabularx}
\label{tb:cls-pop-4}
\end{table*}

\begin{table*}[ht]
\small
\centering
\caption{Comparison of Responses between Cluster 5 and the Overall Population.}
\begin{tabularx}{\textwidth}{XXX}
\toprule
\textbf{Question} & \textbf{Cluster 5 Response} & \textbf{Overall Population Response} \\
\midrule
\textbf{How much pressure, if any, did you feel from family members to marry your partner after you moved in together?} & A lot of pressure: 2.13\% \newline \textbf{Some pressure: 87.26\%} \newline Not too much pressure: 10.59\% \newline  No pressure at all: 0.0\% & A lot of pressure: 21.97\% \newline Some pressure: 10.68\% \newline \textbf{Not too much pressure: 62.09\%} \newline  No pressure at all: 5.24\% \\
\midrule
\textbf{How much control do you think you have over the data the government collects about you?} & A great deal of control: 2.87\% \newline \textbf{Some control: 87.09\%} \newline Very little control: 10.02\% \newline  No control: 0.0\% & A great deal of control: 21.91\% \newline Some control: 11.47\% \newline \textbf{Very little control: 60.06\%} \newline No control: 6.53\% \\
\midrule
\textbf{When you are asked to agree to a company's privacy policy, how often do you read it before agreeing to it?} & Always: 4.19\% \newline \textbf{Often: 87.42\%} \newline Sometimes: 8.38\% \newline Never: 0.0\% & Always: 22.24\% \newline Often: 12.63\% \newline \textbf{Sometimes: 62.87\%} \newline Never: 2.24\% \\
\bottomrule
\end{tabularx}
\label{tb:cls-pop-5}
\end{table*}


\begin{table*}
\small
\centering
\caption{Questions pertaining to crime that exhibit the most disagreement among clusters.}
\begin{tabularx}{\textwidth}{XXX}
\toprule
\textbf{Question} & \textbf{Cluster} & \textbf{Response} \\
\midrule
How often, if ever, do you participate in online discussion forums about guns & Cluster-0 \newline Cluster-1 \newline Cluster-2 \newline Cluster-3 \newline Cluster-4 \newline Cluster-5 & Hardly ever (72.24\%)
\newline Often (77.92\%) 
\newline Hardly ever (77.21\%)
\newline Never (97.01\%) 
\newline Hardly ever (72.74\%) 
\newline Sometimes (84.22\%) \\
\midrule
How important, if at all, is being a gun owner to your overall identity? & Cluster-0 \newline Cluster-1 \newline Cluster-2 \newline Cluster-3 \newline Cluster-4 \newline Cluster-5 & Somewhat important (84.41\%)
\newline Very important (95.35\%) 
\newline Somewhat important (72.45\%) 
\newline Not too important (89.26\%) 
\newline Somewhat important (76.27\%) 
\newline Very important (82.0\%) \\
\bottomrule
\end{tabularx}
\label{tb:gun}
\end{table*}


\begin{table*}[ht]
\small
\centering
\caption{Questions pertaining to race that exhibit the most disagreement among clusters.}
\begin{tabularx}{\textwidth}{XXX}
\toprule
\textbf{Question} & \textbf{Cluster} & \textbf{Response} \\
\midrule
Are the country's current economic conditions helping or hurting people who are Hispanic? & Cluster-0 \newline Cluster-1 \newline Cluster-2 \newline Cluster-3 \newline Cluster-4 \newline Cluster-5 & Helping a little (70.42\%)
\newline Helping a lot (94.54\%)
\newline Hurting a little (59.44\%) 
\newline Hurting a little (65.74\%) 
\newline Hurting a little (42.94\%) 
\newline Helping a little (52.25\%) \\
\midrule
By 2050, a majority of the population will be made up of blacks, Asians, Hispanics, and other racial minorities. In terms of its impact on the country, do you think this will be & Cluster-0 \newline Cluster-1 \newline Cluster-2 \newline Cluster-3 \newline Cluster-4 \newline Cluster-5 & A somewhat bad thing (52.33\%)
\newline A very good thing (97.6\%)
\newline A somewhat good thing (66.44\%)
\newline A somewhat good thing (75.7\%)
\newline Neither a good nor bad thing (48.82\%)
\newline A very good thing (55.46\%) \\
\bottomrule
\end{tabularx}
\label{tb:race}
\end{table*}

\end{document}